\documentclass[letterpaper]{article} % DO NOT CHANGE THIS
\usepackage{aaai24}  % DO NOT CHANGE THIS
\usepackage{times}  % DO NOT CHANGE THIS
\usepackage{helvet}  % DO NOT CHANGE THIS
\usepackage{courier}  % DO NOT CHANGE THIS
\usepackage[hyphens]{url}  % DO NOT CHANGE THIS
\usepackage{graphicx} % DO NOT CHANGE THIS
\usepackage{natbib}  % DO NOT CHANGE THIS AND DO NOT ADD ANY OPTIONS TO IT
\usepackage{caption} % DO NOT CHANGE THIS AND DO NOT ADD ANY OPTIONS TO IT
\usepackage{algorithm}
\usepackage{algorithmic}
\usepackage{booktabs}
\usepackage{siunitx}
\usepackage{multirow}
\usepackage{amsmath}

\usepackage{newfloat}
\usepackage{listings}
\DeclareCaptionStyle{ruled}{labelfont=normalfont,labelsep=colon,strut=off} % DO NOT CHANGE THIS
\floatstyle{ruled}
\newfloat{listing}{tb}{lst}{}
\floatname{listing}{Listing}
\title{Progressive Distillation Based on Masked Generation Feature Method for Knowledge Graph Completion}
\author{
    Cunhang Fan\textsuperscript{\rm 1},
    Yujie Chen\textsuperscript{\rm 1},
    Jun Xue\textsuperscript{\rm 1},
    Yonghui Kong\textsuperscript{\rm 1},
    Jianhua Tao\textsuperscript{\rm 2,\rm 3\thanks{Corresponding authors.}},
    Zhao Lv\textsuperscript{\rm 1*}
}
\affiliations{
    \textsuperscript{\rm 1} Anhui Province Key Laboratory of Multimodal Cognitive Computation, 
    \\ School of Computer Science and Technology, Anhui University\\
    \textsuperscript{\rm 2} Department of Automation, Tsinghua University
    \\ \textsuperscript{\rm 3} Beijing National Research Center for lnformation Science and Technology, Tsinghua University\\
    \{cunhang.fan, kjlz\}@ahu.edu.cn, \{e22201148, e21201068, e22201093\}@stu.ahu.edu.cn, jhtao@tsinghua.edu.cn
}

\usepackage{bibentry}
\begin{document}

\maketitle

\begin{abstract}
In recent years, knowledge graph completion (KGC) models based on pre-trained language model (PLM) have shown promising results. However, the large number of parameters and high computational cost of PLM models pose challenges for their application in downstream tasks. This paper proposes a progressive distillation method based on masked generation features for KGC task, aiming to significantly reduce the complexity of pre-trained models. Specifically, we perform pre-distillation on PLM to obtain high-quality teacher models, and compress the PLM network to obtain multi-grade student models. However, traditional feature distillation suffers from the limitation of having a single representation of information in teacher models. To solve this problem, we propose masked generation of teacher-student features, which contain richer representation information. Furthermore, there is a significant gap in representation ability between teacher and student. Therefore, we design a progressive distillation method to distill student models at each grade level, enabling efficient knowledge transfer from teachers to students. The experimental results demonstrate that the model in the pre-distillation stage surpasses the existing state-of-the-art methods. Furthermore, in the progressive distillation stage, the model significantly reduces the model parameters while maintaining a certain level of performance. Specifically, the model parameters of the lower-grade student model are reduced by 56.7\% compared to the baseline. 
\end{abstract}

\section{Introduction}

Knowledge graphs (KGs) are graph-structured knowledge bases, typically composed of triples $(\mathrm{head}\:\mathrm{entity},\:\mathrm{relation},\:\mathrm{tail}\:\mathrm{entity})$, abbreviate as $(h,\:r,\:t)$. Well known are YAGO \cite{suchanek2007yago}, Freebase \cite{bollacker2008freebase}, Wikidata \cite{vrandevcic2014wikidata} etc. KGs have proved to be useful in various downstream tasks such as intelligent question answering \cite{jia2021complex, saxena2020improving}, recommendation systems \cite{wang2019knowledge, cao2019unifying}, semantic search \cite{xiong2017explicit, berant2014semantic} etc. Despite the significant advances that KGs have made for various applications, they still suffer from the problem of incompleteness as the information in the real world continues to grow. Therefore, for the automatic construction of KGs, knowledge graph completion techniques are crucial.

Existing knowledge graph completion (KGC) tasks can generally be divided into two categories: structure-based and description-based methods. Structure-based methods use the topology and triple structure information of the knowledge graph to represent feature vectors of entity relationships, including TransE\cite{bordes2013translating}, ConvE\cite{dettmers2018convolutional}, and R-GCN\cite{schlichtkrull2018modeling}. Description-based methods, use pre-trained language models and introduce semantic descriptions of entities and relations to learn representations, such as commonly used models like KG-BERT \cite{yao2019kg}, PKGC \cite{lv2022pre}, and LP-BERT \cite{li2022multi}. It is evident that with the rise of pre-trained language models (PLM), description-based methods have gradually taken the lead. By utilizing entity and relation semantic descriptions as auxiliary information and deeply mining the potential knowledge in PLM, description-based methods solve the problem of inductive KGC tasks that structure-based methods cannot handle, while achieving significant improvements in transductive KGC tasks. However, while description-based approaches improve performance, they also bring with them the problems of large model parameter numbers and high computational costs, limiting their application in downstream tasks such as real-time recommendation systems etc. Therefore, model lightweighting is essential.

Knowledge distillation \cite{hinton2015distilling}, which involves the transfer of latent knowledge from a large teacher model to a small student model using soft labels, is a common method for model compression. It has been widely applied in the fields of computer vision \cite{zhao2022decoupled} and speech recognition \cite{kurata2018improved}. In the field of KGC, there are also related research works \cite{zhu2022dualde, wang2021mulde} that employ ensemble models consisting of multiple structure-based KGC models as multi-teacher models to transfer knowledge to student models in order to reduce embedding dimensions. However, to our knowledge, description-based KGC method does not have a corresponding knowledge distillation method, nor can  structure-based KGC distillation strategy be directly applied to the description-based KGC method, because intuitively the model architectures of the description-based KGC method and the structure-based KGC method are too different to be directly migrated. Therefore, we believe that the description-based method needs a simple and efficient knowledge distillation framework to fill this gap.

In this paper, we propose a novel progressive distillation strategy based on masked generation feature (\textbf{PMD}), which can achieve a substantial reduction in model parameters while minimizing the impact on model performance. Traditional feature distillation only learns the representation information of the input entity set. In contrast, masked generation feature distillation potentially learns the representation information of the inferred entity set through inference generation. This approach addresses the problem of single representation information in the teacher model during traditional feature distillation. However, In the case of a limited number of parameters, how the student model can efficiently learn the rich representation information in the teacher model is a problem. To address this problem, we propose an progressive distillation strategy. The strategy aims to enhance inter-model migration efficiency through two approaches: gradually decreasing the mask ratio and reducing the number of model parameters. The objective is to align the amount of mask feature information in the teacher model with the representation capability of the student model. Specifically, we divide the progressive distillation strategy into two stages: in the pre-distillation stage, we use the masked generation feature distillation method to enhance the performance of baseline and to serve as a teacher model to guide the senior student model, and in the progressive distillation stage, we design a multi-grade student model and distills it grade-by-grade. This process focuses on transferring knowledge regarding rich masked generation representations and global triplet information.

We conduct extensive experiments on two representative datasets, and the experimental results demonstrate the effectiveness of PMD. The model in the pre-distillation stage achieves state-of-the-art (SOTA) performance on the WN18RR dataset, the model in the progressive distillation stage can reduce the parameter count by up to 56.7\% compared to baseline while maintaining a certain level of performance. Furthermore, we further validate the significance of masked generation feature distillation and the progressive distillation strategy through ablation experiments.

Our contributions can be summarized as follows:
\begin{itemize}
\item  We propose a progressive distillation strategy based on masked generation features, greatly reducing model complexity and filling the gap in the field of knowledge distillation with description-based KGC methods.
\item We find that the traditional feature distillation strategy suffers from the problem of a single feature representation of the teacher model, so we propose that masked generation feature distillation motivates the teacher model to transfer rich representation information.
\item By conducting extensive experiments on two widely used datasets, WN18RR and FB15K-237, the results show that PMD achieves SOTA performance on the WN18RR dataset. The number of progressive distillation model parameters can be reduced by up to 56.7\% from baseline.
\end{itemize}

\section{Related Work}

\paragraph{Knowledge Graph Completion}
In recent years, KGC method has developed rapidly. The key idea is to map entities and relations in KGs to a continuous vector space as an embedding representation. Among them, structure-based methods focus on representing the feature vectors of a triple through the structural information of the triple itself or the topology of the KGs. For example, TransE \cite{bordes2013translating} models the triple as a relational translation in Euclidean space; Complex \cite{trouillon2016complex} embeds entities and relations in a complex space to deal with asymmetric relations, while RotatE \cite{sun2018rotate} models the triple as a relational rotation in a complex space. With the development of deep learning, CNN-based and GNN-based approaches have been proposed in the industry, such as ConvE \cite{dettmers2018convolutional} and ConvKB \cite{dai2018novel}, which use CNN to capture the local structural information of each triple; SACN \cite{shang2019end} and CompGCN \cite{vashishth2019composition} extract topological information in KGs to represent the triple. With the rise of PLM (BERT \cite{kenton2019bert}, GPT \cite{radford2018improving}, etc.), a large number of PLM-based KGC methods have emerged, such as KG-BERT \cite{yao2019kg}, StAR \cite{wang2021structure}, etc. To represent entity and relational embeddings, they introduce natural language descriptions of entities and relations as auxiliary information to mine the potential entity-relational knowledge in pre-trained language models \cite{petroni2019language}. 

\paragraph{Knowledge Distillation}
Knowledge Distillation (KD) \cite{hinton2015distilling} is one of the most common techniques in model compression and is widely used in computer vision \cite{zhao2022decoupled, yang2022focal} and natural language processing \cite{sun2019patient, sanh2019distilbert, wang2023greekgc}. The core idea of KD is to use the soft label of a large model (teacher model) to guide the learning of a small model (student model). This has the advantage of reducing the computational and storage resource consumption of the model while ensuring performance, thus making the model lighter. In addition to the role of model compression, KD can also improve model performance, and recent studies \cite{abnar2020transferring, kuncoro2019scalable} have found that KD can transfer inductive biases between neural networks.The self-distillation strategy \cite{Pham2022revisting} stands out for its effective knowledge transfer approach, enhancing performance through distillation within its own network. Compared with the traditional logits distillation, masked generation feature distillation method we propose can ensure that the teacher model can transfer knowledge more efficiently, so that students can learn more enriched teacher knowledge. At the same time, combined with the progressive distillation strategy, it can ensure that the model parameters is significantly reduced while maintaining the model performance as much as possible.
\renewcommand{\dblfloatpagefraction}{0.8}
\begin{figure*}
    \centering
    \includegraphics[width=0.86\textwidth]{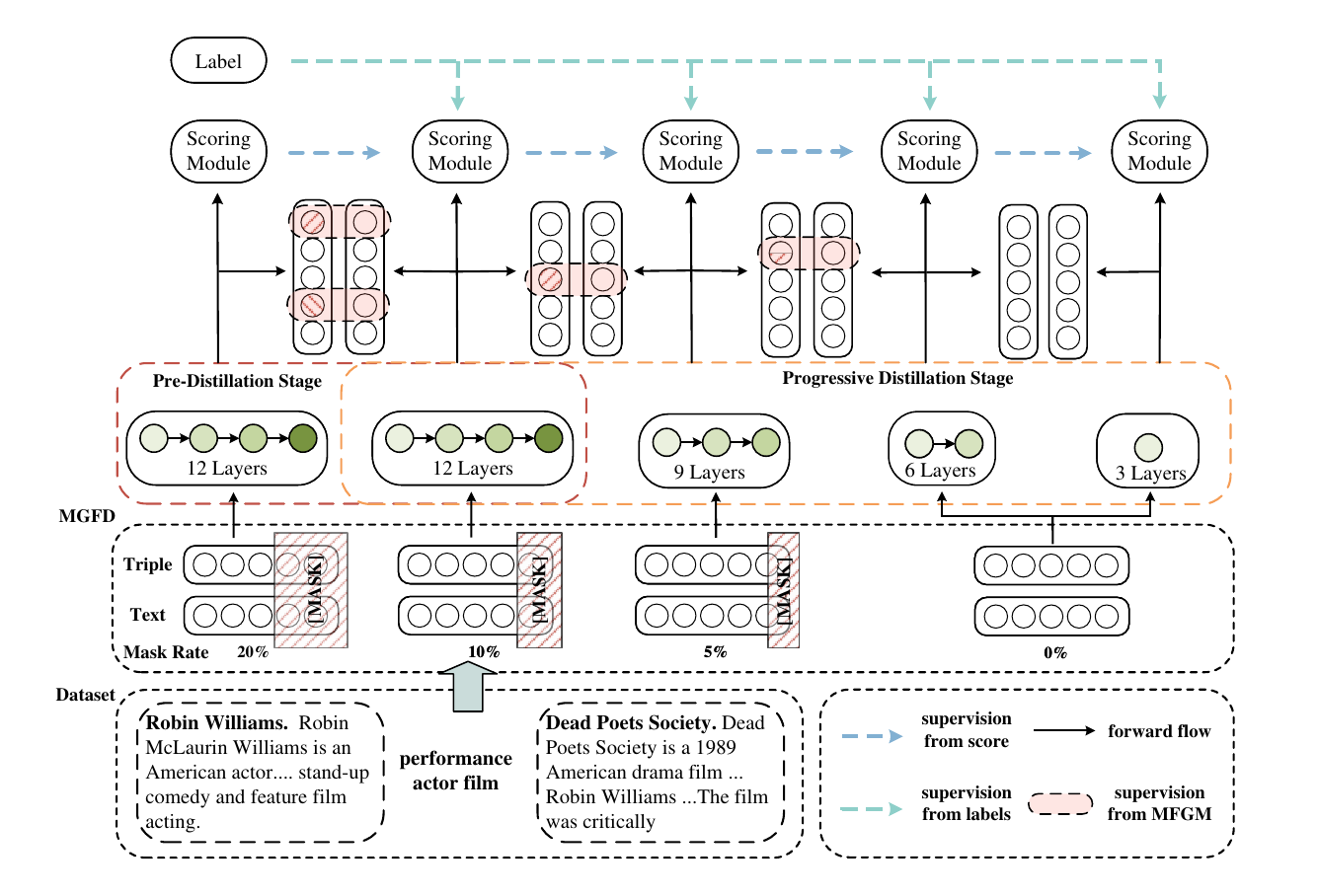}
    \caption{This figure illustrates the overall architecture of the PMD. ($\mathrm{\romannumeral 1}$) MGFD applies masking operations to input tokens and sets an appropriate masking rate based on student model parameter count ($\mathrm{\romannumeral 2}$) In pre-distillation stage, the performance of the initial model is improved. ($\mathrm{\romannumeral 3}$) In progressive distillation Stage involves the design of multi-grade student models with gradually reduced parameter count and mask rate. ($\mathrm{\romannumeral 4}$) Each student model is trained under three kinds of supervision as depicted.}
    \label{fig1:framework}
\end{figure*}

\section{Methodology}
In this section, we introduce PMD in detail, as depicted in Figure.\ref{fig1:framework}. First, we will give a brief overview of knowledge graphs and the definition of link prediction task \ref{Definitions_and_Notation}. Then, we will describe the working principle and implementation details of the masked generation feature distillation \ref{Masked_Generation_Feature_Distillation}. Finally, we will explain the architecture and underlying principles of the progressive distillation framework in detail \ref{Progressive_Distillation_Framework}.

\subsection{Definitions and Notation}
\label{Definitions_and_Notation}
KGs is a directed relational graph consisting of entities and relations. It can be defined as $\mathcal{G} = \{\mathcal{E},\:\mathcal{R},\:\mathcal{T}\}$, where $\mathcal{E}$ is the set of entities, $\mathcal{R}$ is the set of relations and $\mathcal{T}$ is the set of triples, denoted as $T = \{(h,\:r,\:t) \subseteq  \mathcal{E} \times \mathcal{R} \times \mathcal{E}\}$.The link prediction task aims to complete missing triplets based on the existing KGs information. Specifically, under the widely adopted entity ranking evaluation protocol, tail entity prediction infers the tail entity given a head entity and a relation,   head entity prediction is similar. In this paper, inverse triplets are set up for each triplet \cite{dettmers2018convolutional}, so only tail entity prediction needs to be performed in the experiments.

\subsection{Masked Generation Feature Distillation}
\label{Masked_Generation_Feature_Distillation}
BERT \cite{zhang2021inductive} model leverages the technique of masked language modeling to acquire valuable inductive biases. Inspired by this idea, we propose a masked generation feature distillation (MGFD), speculating that the concept of masking can also facilitate the transfer of more inductive biases from the teacher model in knowledge distillation. When triplets and their text description pass through deep networks, they acquire higher-order semantic information. In traditional feature distillation, the semantic information only consists of the global semantic information of neighboring tokens and the semantic information of the current token (i.e.input token set). In contrast, our proposed MGFD not only incorporates the semantic information within the input token set but also utilizes the inferential operations of deep networks to obtain semantic information from the inferred token set. This enrichment leads to a more comprehensive representation of the generated features, enabling the student model to learn more abundant representations. Consequently, this approach addresses the problem of inefficient knowledge transmission from the teacher model in the context of KGC tasks.

Specifically, during the data input stage, we apply a masking operation to the input data of the teacher model and the student model, which includes triplets and their text description, based on a predetermined masking rate determined by the size of the teacher model's parameters. The masked tokens are encoded by the teacher model, resulting in masked feature vectors that encapsulate rich semantic information. Meanwhile, during the training process, the student model encodes the masked tokens and generates corresponding student feature vectors at the masked positions. Ultimately, the two feature vectors are trained using Mean Squared Error (MSE) to progressively align the student's feature vectors with the teacher's. This approach facilitates the transfer of knowledge from the teacher model to the student model, enabling efficient knowledge migration.

Through the above process, the student model learns rich representation information and achieves efficient knowledge transfer from the teacher model. The formula is as follows: 
\begin{equation}
    Mask(\mathcal{F}^{S}\:|\:\lambda) \longrightarrow Mask(\mathcal{F}^{T}\:|\:\lambda)
\end{equation}
Where $\mathcal{F}^{S}$ is the student feature vector, $\mathcal{F}^{T}$ is the teacher feature vector, $\lambda$ is the masking rate of the input sequence, and $Mask$ is a masking operation on part of the input sequence. $\longrightarrow$ denotes the learning process.

\begin{equation}
\label{Loss_MGFD}
    \mathcal{L}_{MGFD}=  MSE(Mask(\mathcal{F}^{S}\:|\:\lambda) - Mask(\mathcal{F}^{T}\:|\:\lambda))
\end{equation}
It should be noted that we only calculate the distillation loss for the masked tokens.

\subsection{Progressive Distillation Framework}
\label{Progressive_Distillation_Framework}
In this section, we present the progressive distillation framework. Specifically, inspired by the idea of layer-by-layer distillation to transfer knowledge from deep models to shallow models \cite{xue2023learning}.By gradually reducing the mask ratio and model parameters, it ensures that teacher model knowledge is effectively transferred to student model.  This solves the issue of mismatched masking-generated feature information from the teacher model and the expressive capacity of the student model.

Specifically, the progressive distillation framework can be divided into two stages: pre-distillation stage and progressive distillation stage. In the pre-distillation stage, we use MGFD to inspire the baseline model's potential and generate a high-quality teacher model. In the progressive distillation stage, we repeatedly compress the teacher model to obtain multi-grade student models. During the distillation process, each higher-grade model is paired with scoring module and MGFD module. The scoring module assigns scores to the triples based on their plausibility, while the MGFD modules gradually reduce the mask rate as the grade decreases.

In the specific training process, PMD performs knowledge transfer through the following key processes. Firstly, it is well known that for most machine learning tasks, truth labels are crucial and contain a large amount of standard information. For the KGC task is no exception, in the process of tail-entity prediction, the matching of correct tail-entity labels often results in high scores in the scoring module, allowing the model to learn the key triple feature information in the dataset. Therefore, true label distillation is essential in the distillation process.

\begin{equation}
    score=cos(e_{hr}, e_{t})=\frac{e_{hr} \cdot e_{t}}{\lVert e_{hr} \rVert \lVert e_{t} \rVert }
\end{equation}
\begin{equation}
    \mathcal{L}_{CE}=CrossEntropy(score, L) 
\end{equation}
where $e_{hr}$ and $e_t$ represent the head entity relationship feature vector and the tail entity feature vector, respectively. $L$ is the true label of the training dataset. $\mathcal{L}_{CE}$ is computed by CrossEntropy.

Compared to the absolute standard information in the true label, the potential prior knowledge in the teacher model is also crucial. The teacher model encodes the global features of the triplets through the PLM model and evaluates the global information of the triplets through the scoring module, which contains much implicit knowledge. Through the MSE function, the student model approaches the evaluation results of the teacher model as closely as possible, which stimulates the learning ability of the student model and enables it to learn the prior knowledge in the teacher model.
\begin{equation}
    \mathcal{L}_{SCORE} = MSE(score^S - score^T)
\end{equation}
where $score^S$ and $score^T$ denote the output of the scoring module for the student and teacher models respectively.

In addition to the information from the triplets in the input sequence, the information of the inferred entity set that matches the triplets is also crucial. The mentioned MGFD in section \ref{Masked_Generation_Feature_Distillation} precisely addresses this key issue. By performing random masking operations, the teacher model generates corresponding inference feature vectors, which contain representation information of the inferred entity set. This compels the student model to learn additional information, thereby enhancing the expressive capability of the student model.The specific formula is given in (\ref{Loss_MGFD}).

Overall, total loss comprises the three components above.  $\alpha$ and $\beta$ are used to balance the model's ability to capture both the global and local information of the triplets, which can be expressed using the following formula:
\begin{equation}
    \mathcal{L} = (1 - \alpha - \beta) * \mathcal{L}_{CE} + \alpha * \mathcal{L}_{SCORE} + \beta * \mathcal{L}_{MGFD}
\end{equation}

\section{Experiments}

\begin{table}
  \small
  \centering
  \begin{tabular}{cccccc}
    \toprule
    Dataset     & $N_{e}$    & $N_{r}$  & $N_{Train}$  & $N_{Valid}$  & $N_{Test}$    \\
    \midrule
    WN18RR      &\num[group-separator={,}]{40943}   &\num[group-separator={,}]{11}   &\num[group-separator={,}]{86835}  &\num[group-separator={,}]{3034}   &\num[group-separator={,}]{3134}     \\
    FB15K-237   &\num[group-separator={,}]{14541}   &\num[group-separator={,}]{237}  &\num[group-separator={,}]{272115} &\num[group-separator={,}]{17535}  &\num[group-separator={,}]{20466} \\
    \bottomrule
  \end{tabular}
  \caption{Statistics of the datasets.}
  \label{datasets}
\end{table}

\subsection{Experimental Setup}

\begin{table*}
\centering
\begin{tabular}{lccccc|cccc}
\toprule
\multicolumn{1}{l}{\multirow{2}{*}{Method}} & \multicolumn{1}{c}{\multirow{2}{*}{P}} & \multicolumn{4}{c}{WN18RR} & \multicolumn{4}{c}{FB15k-237} \\  
\cmidrule(r){3-10}
\multicolumn{1}{c}{} & \multicolumn{1}{c}{}  & MRR   & \multicolumn{1}{l}{H@1} & \multicolumn{1}{l}{H@3} & \multicolumn{1}{l}{H@10} & MRR   & \multicolumn{1}{l}{H@1} & \multicolumn{1}{l}{H@3} & \multicolumn{1}{l}{H@10} \\ 
\midrule
\multicolumn{10}{l}{structure-based   methods}\\ 
\midrule
TransE\cite{bordes2013translating}  & - & 24.3  & 4.3 & 44.1 & 53.2 & 27.9 & 19.8 & 37.6 & 44.1 \\ 
RotatE\cite{sun2018rotate}  & - & 47.6  & 42.8 & 49.2 & 57.1 & 33.8 & 24.1 & 37.5 & 53.3 \\ 
ConvE\cite{dettmers2018convolutional}   & - & 43.0 & 40.0 & 44.0 & 52.0 & 32.5 & 23.7 & 35.6 & 50.1 \\ 
CompGCN\cite{vashishth2019composition} & - & 47.9  & 44.3 & 49.4 & 54.6 & 35.5 & 26.4 & 39.0 & 53.5 \\ 
\midrule
\multicolumn{10}{l}{description-based   methods}\\ 
\midrule
KG-BERT\cite{yao2019kg} & 110M & 21.6  & 4.1 & 30.2 & 52.4 & - & - & - & 42.0 \\
MTL-KGC\cite{kim2020multi} & 110M & 33.1  & 20.3 & 38.3 & 59.7 & 26.7 & 17.2 & 29.8 & 45.8 \\
C-LMKE\cite{Wang2022Language} & 110M & 61.9  & 52.3 & 67.1 & 78.9 & 30.6 & 21.8 & 33.1 & 48.4 \\
KGLM\cite{youn2022kglm}    & 355M & 46.7  & 33.0 & 53.8 & 74.1 & 28.9 & 20.0 & 31.4 & 46.8 \\
LP-BERT\cite{li2022multi}    & 355M & 48.2 & 34.3 & 56.3 & 75.2 & 31.0  & 22.3 & 33.6 & 49.0 \\
StAR\cite{wang2021structure}    & 355M & 40.1  & 24.3 & 49.1 & 70.9 & 29.6  & 20.5 & 32.2 & 48.2 \\ 
$\mathrm{Baseline}$\cite{wang2022simkgc} & 210M & 67.1  & 58.5 & 73.1 & 81.7 & 33.3 & 24.6 & 36.2 & 51.0 \\ 
\midrule \midrule
$\mathrm{PMD_{12}(ours)}$   & \textbf{210M} & \textbf{67.8} & \textbf{58.8} & \textbf{73.7} & \textbf{83.2} & 33.3 & 24.3 & 36.3 & 51.8 \\
$\mathrm{PMD_{9}(ours)}$    & 176M & 67.2 & 58.2 & 73.2 & 82.5 & 32.6 & 23.5 & 35.4 & 51.0 \\ 
$\mathrm{PMD_{6}(ours)}$    & 133M & 65.9 & 56.5 & 72.3 & 81.9 & 32.4 & 23.4 & 35.4 & 50.7 \\
$\mathrm{PMD_{3}(ours)}$    & \textbf{91M} & \textbf{62.8} & \textbf{52.9} & \textbf{69.5} & \textbf{80.4} & 32.3 & 23.3 & 35.2 & 50.5 \\
\bottomrule
\end{tabular}
\caption{Main results for WN18RR and FB15K-237 datasets, "12", "9", and "6" refer to the number of layers in the Transformer Encoder. "H@k" represents "Hits@k". "P" represents Parameters, "M" is short for million.}
\label{table 2}
\end{table*}

\textbf{Datasets.} We experiment on two common KGC benchmark datasets WN18RR\cite{dettmers2018convolutional} and FB15k-237\cite{toutanova2015observed}. WN18RR is a subset of  WordNet\cite{miller1995wordnet}, while FB15K-237 is a subset of Freebase. WN18RR and FB15K-237 resolve the test set leakage problem in WN18 and FB15K by eliminating inverse relations. The statistical data is shown in Table \ref{datasets}.

\begin{table*}[h]
\centering
\setlength\tabcolsep{4pt}
\begin{tabular}{ccccccccccccc}
\toprule
\multirow{2}{*}{\textbf{L}} & \multicolumn{3}{c}{$\mathbf{Baseline^{*}}$}                                & \multicolumn{3}{c}{\textbf{LKD}}                                     & \multicolumn{3}{c}{\textbf{PKD}}                                     & \multicolumn{3}{c}{\textbf{PMD}}            \\ \cmidrule(r){2-13}
                            & MRR & \multicolumn{1}{l}{H@1}              & \multicolumn{1}{l|}{H@10}            & MRR & \multicolumn{1}{l}{H@1}              & \multicolumn{1}{l|}{H@10}   
                            & MRR & \multicolumn{1}{l}{H@1}              & \multicolumn{1}{l|}{H@10}
                            & MRR & \multicolumn{1}{l}{H@1}              & \multicolumn{1}{l}{H@10} \\ \midrule
\textbf{12}                 & 66.9         & \multicolumn{1}{c}{58.4} & \multicolumn{1}{c|}{81.7} & 67.1         & \multicolumn{1}{c}{58.8} & \multicolumn{1}{c|}{81.5} & 67.0         & \multicolumn{1}{c}{58.7} & \multicolumn{1}{c|}{81.2} & 67.8         & 58.8         & \textbf{83.2}          \\
\textbf{9}                  & 63.1         & \multicolumn{1}{c}{53.6} & \multicolumn{1}{c|}{79.5} & 66.2         & \multicolumn{1}{c}{57.5} & \multicolumn{1}{c|}{81.5} & 66.7         & \multicolumn{1}{c}{58.1} & \multicolumn{1}{c|}{81.3} & 67.2         & 58.2         & \textbf{82.5}          \\
\textbf{6}                  & 62.1         & \multicolumn{1}{c}{52.7} & \multicolumn{1}{c|}{78.5} & 64.5         & \multicolumn{1}{c}{55.3} & \multicolumn{1}{c|}{80.8} & 65.4         & \multicolumn{1}{c}{56.4} & \multicolumn{1}{c|}{81.4} & 65.9         & 56.5         & \textbf{81.9}          \\
\textbf{3}                  & 61.2         & \multicolumn{1}{c}{52.4} & \multicolumn{1}{c|}{74.9} & 61.8         & \multicolumn{1}{c}{51.5} & \multicolumn{1}{c|}{79.8} & 62.6         & \multicolumn{1}{c}{52.5} & \multicolumn{1}{c|}{80.1} & 62.8         & 52.9         & \textbf{80.4}          \\ \bottomrule
\end{tabular}
\caption{The comparison experiment of the distillation strategy on the WN18RR dataset, "LKD\cite{hinton2015distilling}" is the logits distillation, and "PKD\cite{sun2019patient}" is the teacher model's middle feature layer, and the feature distillation is performed by layer skipping. “L” is Layer.}
\label{table 3}
\end{table*}

\begin{table*}[]
\centering
\begin{tabular}{ccccccc}
\toprule
\textbf{Method}                                                             
 & \textbf{Layers} & \textbf{MR} $\downarrow$ & \textbf{MRR} $\uparrow$ 
 & \textbf{Hits@1} $\uparrow$ & \textbf{Hits@3} $\uparrow$ & \textbf{Hits@10} $\uparrow$             \\  
 \midrule
 & 12 & 132.1 & 67.0 & 58.4 & 72.7 & 81.7 \\
 & 9 & 140.0 & 63.1 & 53.6 & 69.6 & 79.5 \\
 & 6 & 130.7 & 62.1 & 52.7 & 68.5 & 78.5 \\
 \multirow{-4}{*}{\textbf{\begin{tabular}[c]{@{}c@{}}$\mathbf{Baseline^{*}}$\\\end{tabular}}} & 3 & 244.3 & 61.2 & 52.4 & 65.8 & 74.9 \\
 \midrule
 & 12     & 145.6 & 67.3 & \textbf{59.3} & 72.5 & 81.3 \\
 & 9      & 150.4 & 66.7 & 58.6 & 72.1 & 80.6 \\
 & 6      & 166.9 & 65.4   & 56.9 & 70.6 & 80.3 \\
 \multirow{-4}{*}{\textbf{\begin{tabular}[c]{@{}c@{}}PMD\\ (w/o MGFD) \end{tabular}}} & 3  & 165.1  & 62.3 & 52.8 & 68.4 & 78.9 \\
 \midrule
 & 12     & \textbf{110.3} & \textbf{67.8} & \underline{58.8} & \textbf{73.7} & \textbf{83.2} \\
 & 9      & 107.2 & 67.2 & 58.2 & 73.2 & 82.5 \\
 & 6      & 120.0   & 65.9 & 56.5 & 72.3 & 81.9 \\
 \multirow{-4}{*}{\textbf{\begin{tabular}[c]{@{}c@{}}PMD\\ (ours) \end{tabular}}} & 3  & 133.6  & 62.8 & 52.9 & 69.5 & 80.4 \\           \bottomrule
 \end{tabular}
 \caption{Main results with and without (w/o) the MGFD module on the WN18RR dataset. $\downarrow$ indicates that the lower the indicator, the better the performance. $\uparrow$ indicates that the higher the indicator, the better the performance}
\label{table 4}
\end{table*}

\textbf{Baselines.} We select a representative SimKGC \cite{wang2022simkgc}  model as the baseline for our distillation framework. The effectiveness of our framework migration is demonstrated by achieving performance breakthroughs on hard-to-breakthrough high metric models.

\textbf{Evaluation Metrics.} For tail entity prediction, given an $(h,\:r,\:?)$ pair, we predict and rank all possible entities and obtain the rank of $t$. The head entity prediction experiment is similar. We use four automatic evaluation metrics: (1) MRR (Mean Reciprocal Rank), the average inverse rank of the test triples. (2) Hits@$k$ ($k \in \{1, 3, 10\}$), the proportion of correct entities ranked in the top k. Higher MRR and Hits@$k$ values indicate better performance. 

\textbf{Hyperparameters.} The encoders of PLM are initialized as \textit{bert-base-uncased},  During the distillation process,The number of layers in the student model is [12, 9, 6, 3] respectively, The masking rates are [20\%, 10\%, 5\%, 0\%] respectively. The weight values of the loss function, $\alpha$ and $\beta$, are searched in a grid with intervals of 0.05 within the range of [0, 0.5]. We perform a grid search on the learning rate range $\{3 \times 10^{-5},\:5 \times 10^{-5}\}$. We use the AdamW optimizer with linear learning rate decay. The model is trained in batch size of 512 on 2 RTX 3090 GPUs. The code of our method has been released in \url{https://github.com/cyjie429/PMD}

\subsection{Main Result}
We reuse the data results from StAR\cite{wang2021structure} regarding TransE and obtained the experimental data for other models from their respective papers' best results. In Table \ref{table 2}, on the WN18RR dataset, the $\mathrm{PMD_{12}}$ during the pre-distillation stage achieve a significant improvement in all metrics without increasing the model parameters, reaching the SOTA level. We attribute this to the MGFD module, which allows the student model to learn rich representation information, thereby enhancing the model's robustness and significantly improving its overall performance.

For the FB15K-237 dataset, $\mathrm{PMD_{12}}$ shows improvements in Hits@3 and Hits@10 metrics, but there is a decrease in Hits@1. We believe this is mainly due to two reasons. Firstly, on the FB15K-237 dataset, there are only 14,541 entities and 237 relations, with an average in-degree of 37 for entities. This implies that an entity can correspond to multiple relations. Therefore, in the MGFD, masking triples may lead to incorrect inferences by the teacher model due to the multitude of relationships, resulting in what is known as teacher giving wrong answers. Consequently, the student model learns incorrect representation information, leading to a decrease in Hits@1. Secondly, the baseline model's performance on FB15K237 is not satisfactory, which means that the teacher model did not learn strong inductive biases. This slight issue of erroneous transfer occurs as a result. 

In Table \ref{table 2}, we compare the parameters of popular PLM-based KGC methods. From the experimental results, we observe the following. $\mathrm{PMD_{9}}$ surpasses the baseline in all metrics except for Hits@1, even with a reduced parameter count. This indicates that the PMD method can ensure model performance while achieving model network compression. Then, $\mathrm{PMD_{3}}$  achieves better expressive power compared to commonly used high-parameter models (i.e. 110M, 355M) with a reduced model network size of only 91M (i.e. 56.7\% reduction compare to the baseline). This demonstrates the effectiveness of our proposed PMD strategy in maintaining good performance despite a significant reduction in model parameters. The difference in parameters still remains substantial between the structure-based methods and the description-based methods. However, the two approaches address distinct problems. Description-based methods, employing pre-trained language models, can tackle inductive KGC tasks, which involve predicting unseen entities. In contrast, structure-based methods are confined to performing KGC tasks within known entity sets. 

\section{Ablation}
\subsection{Q1: Is PMD More Efficient Than Common Distillation Strategies?}
\textbf{Yes!} We choose a common and powerful distillation strategy to do a comparative experiment, the specific experiment is shown in Table \ref{table 3}. From Table \ref{table 3}, it can be found that because the representation capacity of the model is constrained by the number of model parameters, when the number of model parameters decreases sharply, the model's performance will also decline. However, knowledge distillation methods can mitigate this performance degradation to some extent. By comparing LKD, PKD and PMD, we find that the use of knowledge distillation methods can greatly alleviate the substantial drops in Hits@1 and Hits@10 metrics. 

However, when comparing LKD, PKD, and PMD, there still exists a noticeable gap, particularly in terms of the Hits@10 metric. Comparing the 12-layer models in the pre-distillation stage, only our PMD achieves performance improvement across all metrics. This allows the model to surpass its own capabilities without increasing the model's workload. We believe that the main reason is the effectiveness of the MGFD module. In comparison to the original training strategy, MGFD introduces a greater amount of uncertain reasoning information through masked inference, which includes both positive and negative information. However, because the teacher model is a well-trained strong model, it tends to lean towards the positive side when it comes to reasoning information. This also enables the student model to learn richer knowledge, thereby enhancing the robustness of the model.

\begin{figure*}
    \centering
    \includegraphics[width=\textwidth]{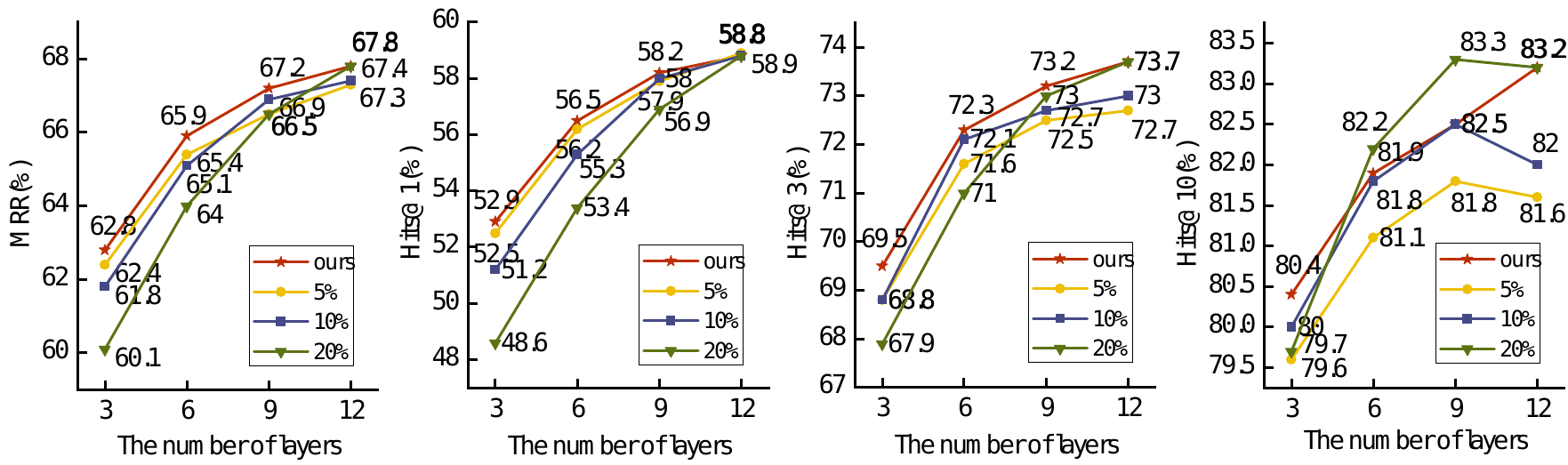}
    \caption{Comparison experiments between the diminishing mask rate and the fixed mask rate.}
    \label{fig2}
\end{figure*}

\subsection{Q2: Are Both Progressive Distillation Module and MGFD Module Useful?}
\textbf{Yes!} Specifically, the difference between PMD (w/o MGFD) and PMD(MGFD) lies in whether or not to perform a mask operation on the input sequence, and the rest of the implementation process is exactly the same. 

As shown in Table \ref{table 4}, when we remove the MGFD module and use only the progressive distillation strategy, the model achieves an improvement in precision on the Hits@1 metric, while the Hits@3 and Hits@10 metrics experience a certain degree of decline. This indicates that the progressive distillation strategy effectively transfers the global representation information from the teacher model to the student model. However, it also introduces the biases present in the teacher model, leading to a decrease in the model's robustness. Moreover, when compare to the baseline model, PMD (without MGFD) outperforms the baseline on most metrics while greatly preserving the performance of the teacher model. This further validates the feasibility and effectiveness of the progressive distillation strategy.

On the other hand, when the MGFD module is employed, taking $\mathrm{PMD_{12}}$ as an example, compared to $\mathrm{PMD_{12}(w/o MGFD)}$, sacrificing a mere 0.4 on the Hits@1 metric results in performance gains of 1.3 on Hits@3 and 1.9 on Hits@10. We believe that such a trade-off is highly valuable, and the same trend holds when the number of layers changes. It addresses the issue of reduced model robustness when using only the progressive distillation strategy. Therefore, we believe that by striking a balance between the two, we can improve the overall performance of the model.

We compare the results of fixed masking rate and decreasing masking rate in our experiment, as shown in Figure \ref{fig2}. We find that using a decreasing masking rate is the most stable and consistently maintains the best performance across all metrics. The fixed 20\% mask rate performs well on the hits@10 metric, but drops dramatically on the rest of the metrics. This demonstrates that only a progressive distillation strategy with simultaneous decreases in mask rate and model parameters can reduce information loss in the teacher model and thus ensure a certain level of model performance while the model parameters plummet.

\subsection{Q3: What Effects Do Different Mask Rates in the MGFD Module Have?}
\begin{figure}
    \centering
    \includegraphics[width=1.0\columnwidth]{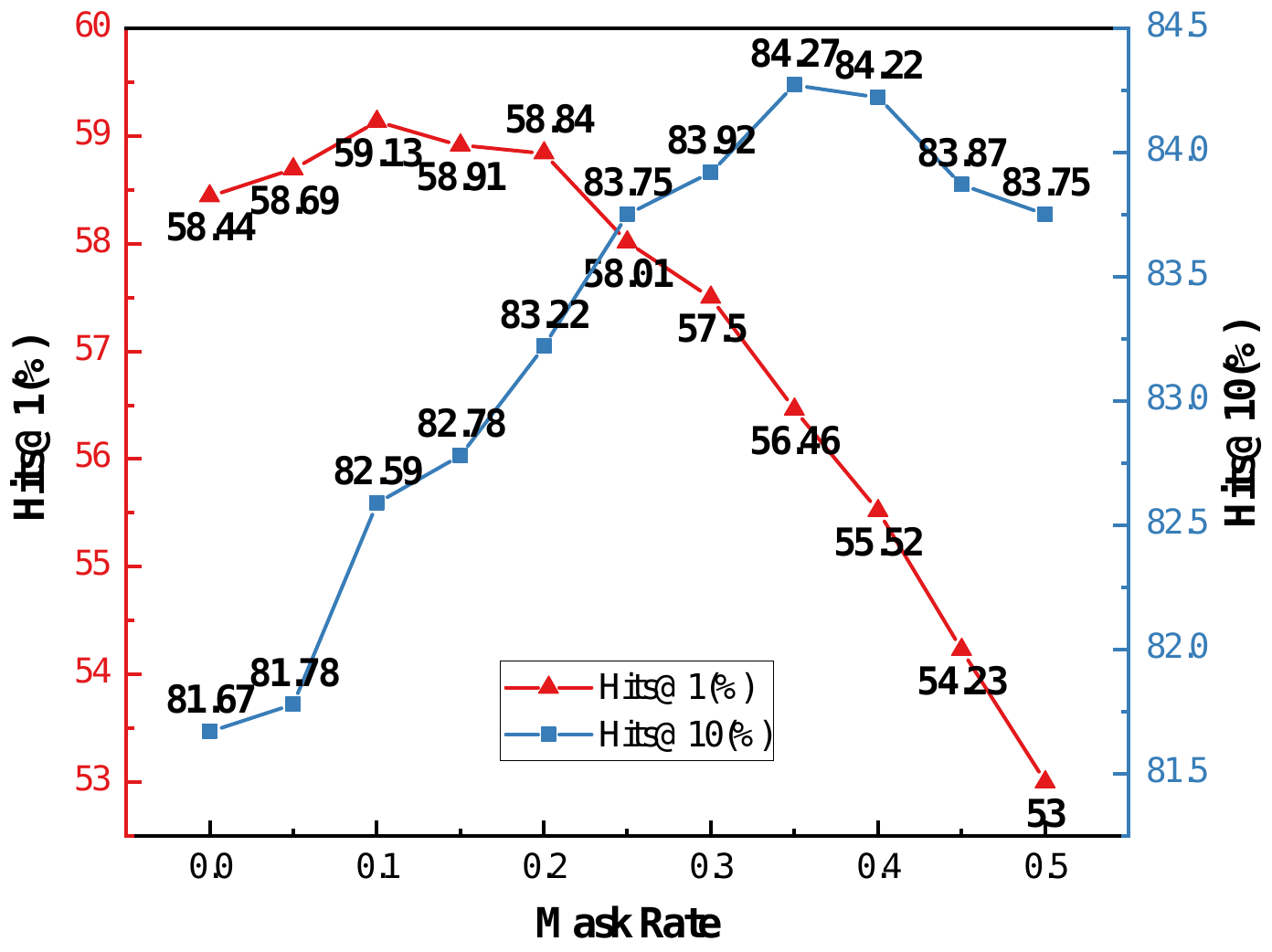}
    \caption{Hits@1 and Hits@10 indicators of the $\mathrm{PMD_{12}}$ with mask rates from 0\% to 50\%.}
    \label{fig3}
\end{figure} 

\textbf{The higher the masking rate, the stronger the model's robustness, but the lower its accuracy.} We only explored the masking rate of MGFD in the pre-distillation stage for the $\mathrm{PMD_{12}}$ model, without increasing the model's parameters, in order to eliminate any performance degradation resulting from reducing the model's parameters. The experimental results, as shown in Figure.\ref{fig3}, As the masking rate gradually increases from 0\% to 50\%, the Hits@10 metric shows a progressive upward trend (i.e., increasing from 81.67 to 84.27 and then decreasing to 83.75), while the Hits@1 metric exhibits a gradual downward trend (i.e., decreasing from 58.44 to 53). This indicates that during the knowledge transfer process, the MGFD module enables the student model to learn rich representations of masked features generated by the teacher model, thereby improving the model's robustness. However, excessively high masking rates can prevent the teacher model from generating accurate triple features based on existing information, resulting in a decrease in precision. Therefore, based on the experimental results, we finally consider a 20\% masking rate as optimal, as it ensures both model robustness and precision improvement.

\section{Conclusion}
In this paper, we propose PMD method, aiming to significantly reduce the complexity of KGC models. To address the issue of limited representation information in traditional feature distillation methods, we design the MGFD approach, where rich representation information is generated by the teacher model and transferred to the student model. To tackle the problem of mismatched expressive power between the teacher and student models, we introduce a progressive distillation strategy that gradually reduces the masking ratio and model parameters, enabling efficient knowledge transfer between teacher and student. Extensive experimental results and ablation studies validate the effectiveness of PMD. In the future, to more effectively transfer knowledge from the teacher model to the student model, we will explore adaptive selection of mask rate and adaptive selection of mask positions during the distillation stage.

\appendix
\section{Acknowledgments}
This work is supported by the {STI 2030—Major Projects (No. 2021ZD0201500)}, the National Natural Science Foundation of China (NSFC) (No.62201002), Distinguished Youth Foundation of Anhui Scientific Committee (No. 2208085J05), Special Fund for Key Program of Science and Technology of Anhui Province (No. 202203a07020008), Open Fund of Key Laboratory of Flight Techniques and Flight Safety, CACC (No, FZ2022KF15)

\bibliography{aaai24}

\end{document}